\documentclass[conference]{IEEEtran}
\usepackage{times} 

\usepackage[sort&compress,numbers]{natbib}
\usepackage{multicol}
\usepackage[bookmarks=true]{hyperref}

\makeatother
\usepackage{amsfonts}

\usepackage{mathtools}
\usepackage{tabularx}
\usepackage{graphicx}
\usepackage{subfigure}
\usepackage{enumerate}
\usepackage{float}
\usepackage{url}
\usepackage{verbatim}
\usepackage{array}
\usepackage{multirow}
\usepackage{longtable}
\usepackage{rotating}
\usepackage{caption}
\usepackage{amssymb}
\usepackage{booktabs}
\usepackage{wrapfig}
\usepackage{adjustbox}
\usepackage{capt-of}

\usepackage{algorithm}
\usepackage{algorithmic}
\usepackage{arydshln}
\usepackage{xcolor}
\usepackage{amsthm}
\usepackage[thicklines]{cancel}
\usepackage{ulem}
\renewcommand{\thetable}{\arabic{table}}
\makeatletter
\newcommand{\setword}[2]{\textbf{#1}\def\@currentlabel{#1}\label{#2}}
\makeatother

\begin{document}

\title{Whole-Body Control Through Narrow Gaps From Pixels To Action}


\author{\authorblockN{Tianyue Wu, Yeke Chen, Tianyang Chen, Guangyu Zhao, and Fei Gao}
	\authorblockA{Institute of Cyber-Systems and Control, Zhejiang University\\ Email: \{tianyueh8erobot, fgaoaa\}@zju.edu.cn}}


%

\maketitle

\begin{abstract}
	Flying through body-size narrow gaps in the environment is one of the most challenging moments for an underactuated multirotor. We explore a purely data-driven method to master this flight skill in simulation, where a neural network directly maps pixels and proprioception to continuous low-level control commands. This learned policy enables whole-body control through gaps with different geometries demanding sharp attitude changes (e.g., near-vertical roll angle). The policy is achieved by successive model-free reinforcement learning (RL) and online observation space distillation. The RL policy receives (virtual) point clouds of the gaps' edges for scalable simulation and is then distilled into the high-dimensional pixel space. However, this flight skill is fundamentally expensive to learn by exploring due to restricted feasible solution space. We propose to reset the agent as states on the trajectories by a model-based trajectory optimizer to alleviate this problem. The presented training pipeline is compared with baseline methods, and ablation studies are conducted to identify the key ingredients of our method. The immediate next step is to scale up the variation of gap sizes and geometries in anticipation of emergent policies and demonstrate the sim-to-real transformation.
\end{abstract}
\vspace{-0.1cm}

\IEEEpeerreviewmaketitle

\section{Introduction}
\label{sec:intro}
Cutting through narrow gaps is one of the most striking moments for flying robots as they traverse the environment. Mastering this motor skill is one of the fundamental requirements for a flying robot to survive in scenarios with very restricted local free space. Nevertheless, these body-sized gaps demand the robot to exploit its asymmetrical vehicle shape and decide the right pose and timing to fly through with low fault tolerance.  This challenge is exacerbated when our flying robots, such as the most common quadrotors in everyday life, are underactuated. While these robots can fly with impressive agility, the control of rotation and translation components in the vehicles are tightly coupled. In this way, the feasible action space can become very limited when approaching a narrow gap and careful action sequences are required when both approaching and traversing through the gap. 

In the past, researchers have utilized trajectory optimization (TO), predefined gap geometry \cite{mellinger2012trajectory, loianno2016estimation, wang2022geometrically, falanga2017aggressive} or prior known gap state \cite{mellinger2012trajectory, loianno2016estimation, wang2022geometrically}, and near-perfect state estimation from motion capture systems  \cite{mellinger2012trajectory,wang2022geometrically} or pre-defined visual features on the gap instance \cite{falanga2017aggressive} to address this challenge. However, when confronted with gaps of unknown geometries and using cheap online sensing, one has to compromise on the size and pose of the gap for safe deployment when using such a modular architecture. This is primarily attributed to information loss across modules and compounding error due to artificially defined inter-module interfaces and isolation optimization objects of each module \cite{bojarski2016end,hu2023planning}. For instance, in such a framework, state estimates generated by an estimator, which is far from perfect with cheap sensors \cite{qin2018vins}, serve as an artificial interface between modules and inevitably become a lossy compression of the raw sensory signals. In this way, the decision-making module is imposed to either pessimistically model uncertainty on \cite{janson2017monte, yang2023safe} or just trust in \cite{ren2023online} the estimates to release actions without access to the sensory signals. Specifically on the task of traversing gaps, a typical workflow can be a TO module to plan a feasible trajectory by trusting in the state estimates, and a controller tuned to blindly minimizes the trajectory tracking error \cite{song2023reaching} according to the state estimates, with no awareness of the task at hand and error of state estimation. Then the practitioners conduct extensive re-engineering, such as narrowing the perceived free space to set an overly conservative 'safe margins' for TO. Unfortunately, this scheme is likely to be compromising, e.g., making it impossible to find a feasible solution for TO once the free space becomes restricted.  

\begin{figure}[t] \centering 
	\vspace{0.2cm}
	\includegraphics[width=0.492\columnwidth]{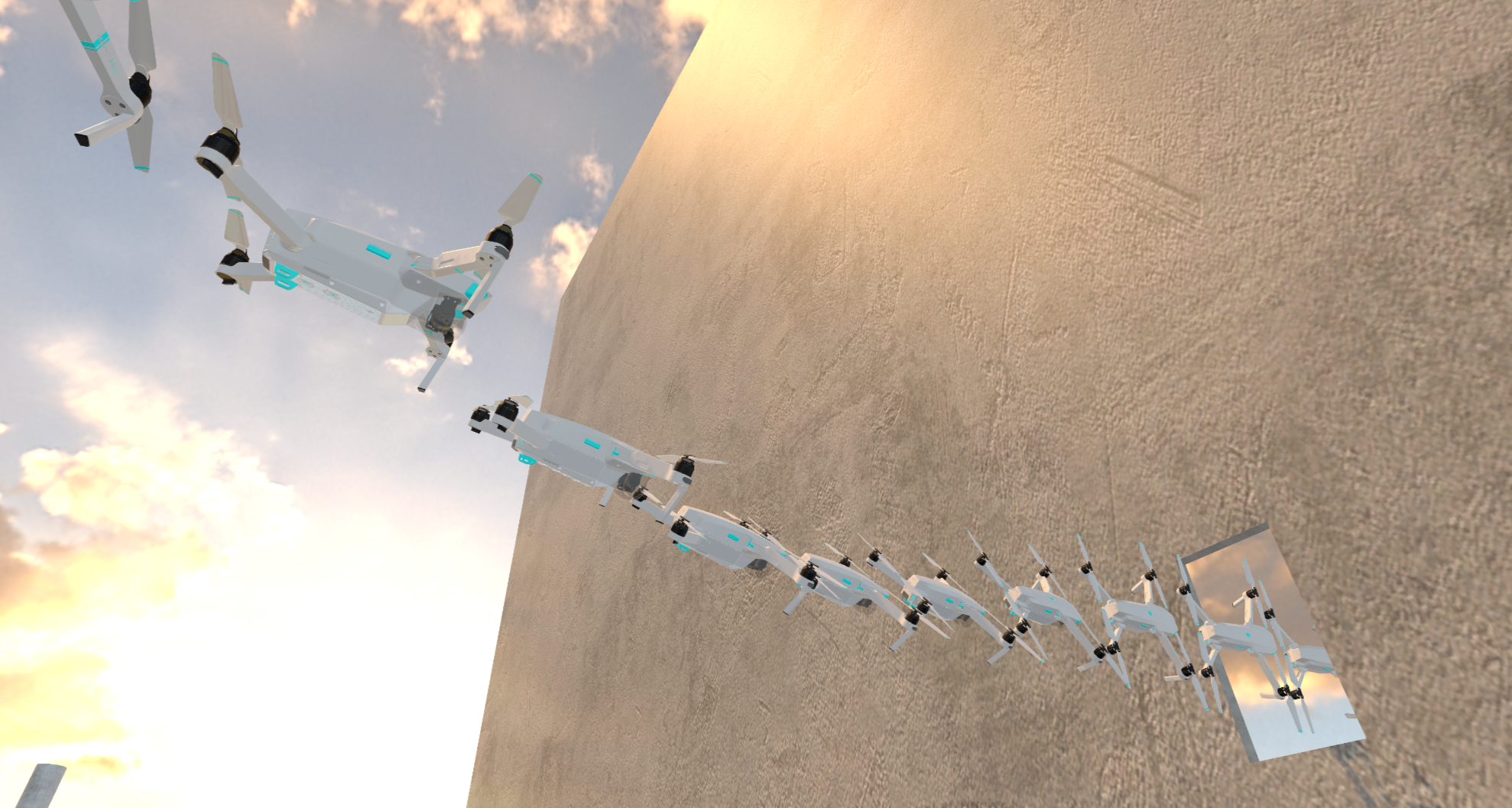} 
	\includegraphics[width=0.492\columnwidth]{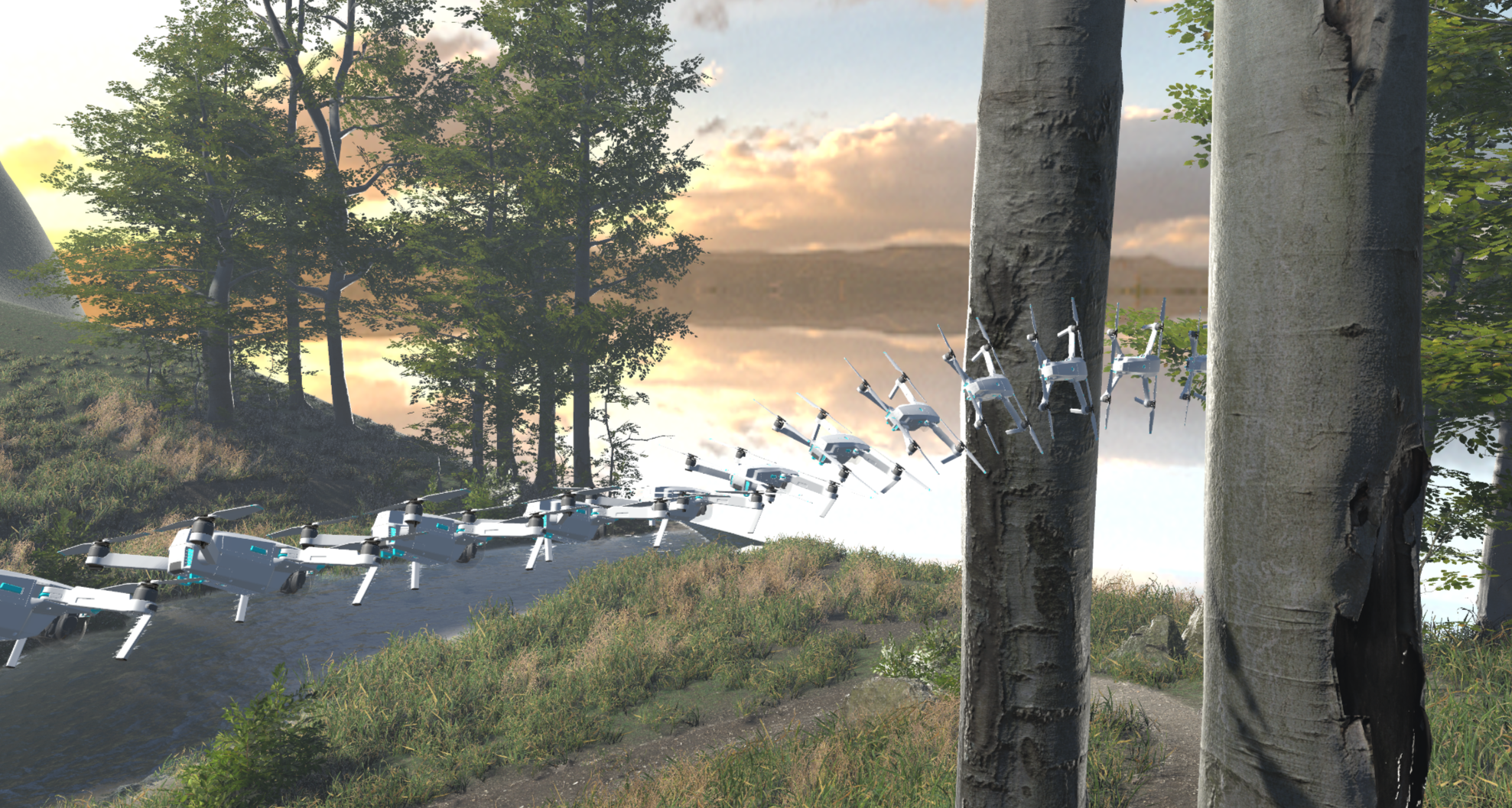}
	\captionsetup{skip=5pt} 
	\captionsetup{font=footnotesize}  
	\caption{\textbf{The snapshots of the quadrotor driven by the pixel-based policy in simulation.} We choose a wall with a hole (i.e., the example in Fig. \ref{fig:method}) and two trees' trunks as masks (Sec. \ref{sec:distillation}), respectively, for the above contexts.}
	\label{fig:head} \vspace{-0.6cm}     
\end{figure} 

End-to-end data-driven methods are pushing the boundaries of solving major challenges that have resisted the most advanced efforts in the robotics community \cite{pan2017agile, hwangbo2019learning, lee2020learning,wurman2022outracing, chen2023visual, kaufmann2023champion, song2023reaching,brohan2023rt, haarnoja2024learning}. They do not require the interfaces between modules to be explicitly defined, \emph{learning} to discover the cues for decision making from sensory data, and optimize the policy in an end-to-end manner \cite{lecun2015deep,amos2018differentiable}. We explore to \emph{use data all the way} to achieve the extreme motor skill of flying through body-size narrow gaps with a \emph{simulated} (underactuated) quadrotor. The policy is designed to directly map observations from onboard sensors, e.g., a camera and an inertial measurement unit (IMU), to the collective thrust and body rates. Specially, we adopt online reinforcement learning (RL) using an ego-centric low-dimensional disordered point representation of the gap to surrogate the pixel observation for scalable data generation in simulation. The training task consists of 5 gaps with different geometries,  random gap poses and the initial states of the agent. Then we transfer the knowledge in the RL policy by aligning the outputs of a recurrent pixel-based policy with those of the RL policy via online imitation learning (IL), which suppresses the covariate shift problem of offline learning \cite{ross2011reduction}. This process is called \emph{distillation} and is widely adopted in policies learning from pixels \cite{chen2023visual,miki2022learning,agarwal2023legged,song2023learning,xu2023unidexgrasp,zhuang2023robot} to allow RL policy to achieve the best performance and to ease the burden of image rendering thanks to the data efficiency of IL.

Unfortunately, in parallel to the difficulty the task poses on model-based techniques, it also causes exploration problems of online RL due to the restricted feasible solution space. The long evolution of classical model-based techniques such as TO \cite{wang2022geometrically} has allowed us to obtain efficient solutions to the problem at hand under perfect self and target state estimation that is availiable in simulation. Even if these solutions are suboptimal due to the nonconvex landscapes of the optimization problems and the decoupled framework of planning and control, they offer informed guidance for the agent to explore. We propose to use the states in the rollout of these classical methods to reset the agent when it triggers termination conditions. This approach is demonstrated to help the agent, without any handcrafted curriculum, to domainate tasks that can not be effectively explored or take much more iterations to learn when using a random reset scheme. This approach, compared to those directly mimic \cite{bojarski2016end,pan2017agile} or synthesize \cite{levine2020offline,chebotar2021actionable} the 'demonstrations', is expected to better leverage online RL's ability to explore and exploit data generated at scale. 

At this stage, the results, albeit very promising, are only verified in noiseless or low-noise simulations and cannot be directly transferred to the real world. We conclude the article by discussing the potential of sim-to-real transformation where we recognize the challenges that may be encountered. In summary, this work
\begin{itemize}
	\item presents the first pixel-to-action policy learning method for the extreme motor skill of flight through narrow gaps;
	\item resets the agent as rollout states by model-based techniques during online RL to alleviate the exploration challenge without elaboratly designed curriculums;
	\item conducts comparision with baseline methods and ablation experiments to identify some of the key technical choices.  
\end{itemize}

\section{Related Work}
\subsection{End-to-end Policy Learning from Pixels for Mobile Robots}
It has been a vision of beauty of the robotics community to realize agile mobile robots that perform closed-loop control in a similar way to humans' driving behavior or animals that make decisions without performing explicit position feedback and trajectory optimization.  

In \cite{pan2017agile}, the authors distill the performance of model predictive control (MPC) into an end-to-end policy network, which directly maps RGB images to low-level control, for off-road car racing. In \cite{loquercio2018dronet}, the authors use a network to drive a drone by mapping the raw RGB images to velocity commands in the horizontal plane without position-feedback control. In \cite{sadeghi2016cad2rl}, the authors propose to use RL policy trained with highly randomized simulated images to control a drone's velocity. Much of the work on legged locomotion present end-to-end pixel-to-action policy to walk on various terrains or avoid obstacles \cite{miki2022learning,yang2021learning,agarwal2023legged,zhuang2023robot,cheng2024extreme,he2024agile}, a large portion of it which is accomplished by means of online RL followed by observation space distillation \cite{miki2022learning,agarwal2023legged,zhuang2023robot,cheng2024extreme}. The authors in \cite{song2023learning} demonstrate end-to-end high-speed obstacle avoidance policy in a prior known environment using RL and distillation.  End-to-end vision-based swarm flight without inter-robot communication is achieved in \cite{zhang2024back} using the differentiable simulator-based policy learning that optimizes a differentiable cost \cite{qiao2020scalable}. This method demonstrates better efficiency than the data-hungry on-policy RL algorithm \cite{schulman2017proximal}. Some works employ online imitation learning (IL) or RL from image abstractions such as gate inner edges or corners \cite{geles2024demonstrating,xing2024bootstrapping} for drone racing. We note that, unlike the drone racing task that seeks extreme speed in fixed track and obstacle avoidance task where the free space is relatively large, flight through body-size gaps requires exploiting full-body control capability instead of using the point model in the above work, and searching a far more restricted feasible solution space, posing unique challenges to the algorithmic design of online RL. Moreover, the poses and geometries of the gaps are not fixed, unlike the gates in drone racing, so we need to handle a larger task variation.   

\subsection{Flight Through Gaps with Underactuated Multirotors}
Flight through narrow gaps is a classic challenge since quadrotors could fly autonomously \cite{mellinger2012trajectory, falanga2017aggressive,loianno2016estimation,wang2022geometrically}. In \cite{mellinger2012trajectory,loianno2016estimation}, TO is performed by pre-determining the \emph{traversal state} with prior knowledge of the gap pose and position. The difference is that the authors of \cite{mellinger2012trajectory} employ a motion capture (mocap) system, while in \cite{loianno2016estimation} a visual-inertial odometry (VIO) from a downward-looking camera is used as explicit state feedback. In \cite{wang2022geometrically}, the authors adopt a more generic abstraction of free space, i.e., polytopes \cite{deits2015computing,wang2024fast}, instead of the pre-determined traversal state, to efficiently solve for a better solution through narrow gaps under a mocap. 
The trajectory is pre-generated without replanning and to address the inaccurate prior knowledge of the gap state, the authors compromised by using a gap considerably larger than the vehicle's size. In contrast to the above efforts, the authors in \cite{falanga2017aggressive} adopt a predefined-geometry gap to define visual keypoints for online state estimation of self and gap states and demonstrate impressive self-closed loop performance. However, they design a specialized trajectory planning and replanning method for rectangle-shaped gaps, instead of applying a generic TO method as in \cite{wang2022geometrically}. Some recent work also adopts online RL or IL with full state from mocap as inputs for flight through fixed-geometry gap \cite{lin2019flying,xiao2021flying,chen2022learning} using pre-determined traversal state. These works do not show, even in simulation, traversals with large attitudes or very narrow spaces like the ones demonstrated in our work. Importantly, our work aims to use high-dimensional pixel information provided by onboard sensors to handle gaps with different geometries and poses without pre-determined traversal states. Moreover, while previous work \cite{xiao2021flying} shows successful sim-to-real transformation, it requires a well-designed handcrafted curriculum during training in simulation. The reset scheme used in this paper enables the agent to directly master the skill of flight through narrower gaps without curriculum. \vspace{-0.1cm}

\section{Problem Statement: Whole-Body Control Through a Gap}
\label{sec:problem}
\begin{figure} \centering 
	\vspace{-0.2cm}
	\hspace{-0.3cm}  
	\includegraphics[width=0.75\columnwidth]{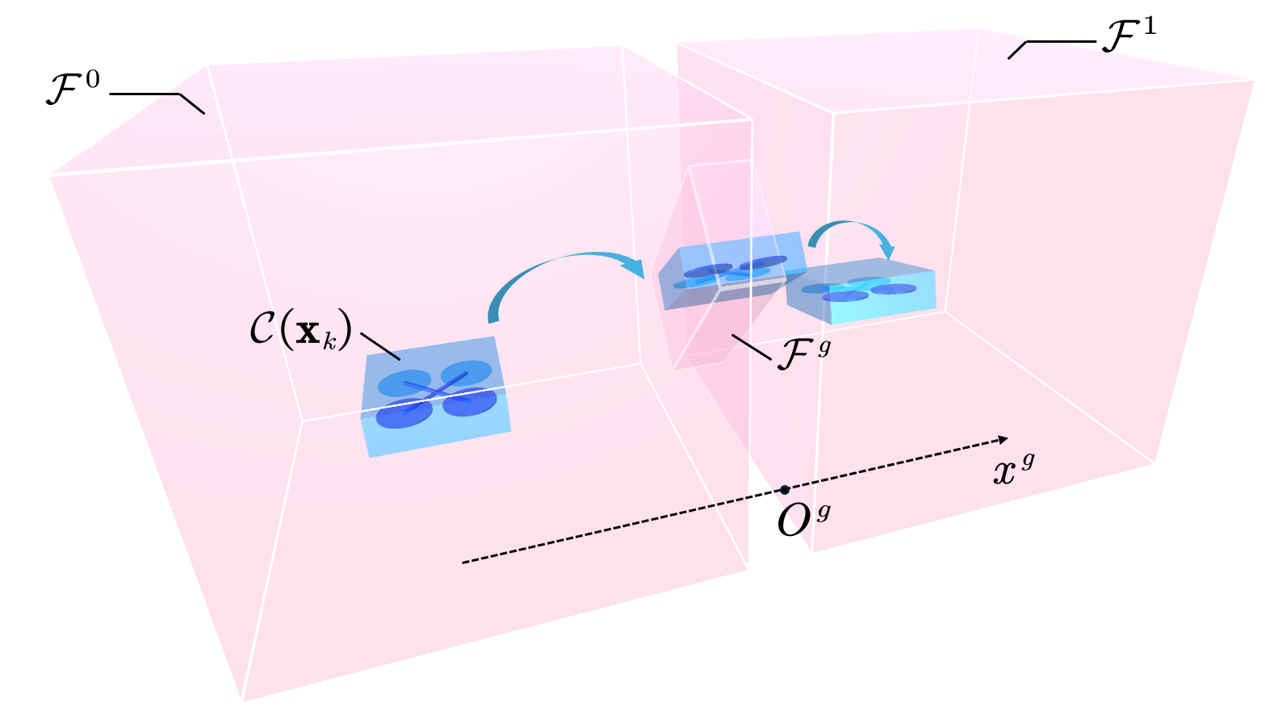} 
	\captionsetup{skip=5pt} 
	\captionsetup{font=footnotesize}  
	\caption{\textbf{Illustration of the problem of whole-body control through a gap.}} \vspace{-0.6cm}  
	\label{fig:illustration}   
\end{figure}

\vspace{-0.1cm}
In this section we formulate the control problem at hand. The vehicle of the (flying) robot is modeled as a rigid body with simplified solid geometry, typically a cuboid or ellipsoid, which is called \emph{collider} for determining if any collision happens. The robot is considered a discrete-time dynamical system with continuous states and control inputs, where the system's state space is defined as the one of its rigid-body vehicle, i.e., $\mathbf{x}_k\in \mathrm{SE}\left( 3 \right)$, $k$ is the time instance. The goal is to find the optimal control sequence to drive the robot through a  plane without collision, on which a closed region called \emph{gap} is the only free space on the plane.  To cover the practical situation where the gap has a thickness, we assume that the plane has a fixed thickness $d$ (set as 10 $\mathrm{cm}$ in this work). Accordingly, we divided the free space into three separate regions, i.e., the gap $\mathcal{F}^g$, the space $\mathcal{F}^0$ in which the robot initially locates, and the other side w.r.t. the plane, $\mathcal{F}^1$. The above geometric relations are visualized in Fig. \ref{fig:illustration}. 

Since the gap can be very narrow, the robot has to adjust its attitude thus utilizing the asymmetry body to traverse through the gap. Therefore, the problem to solve is called the \emph{whole-body control} problem. By writing the geometry corresponding to the collider as a function of its state $\mathbf{x}_k$, $\mathcal{C} \left( \mathbf{x}_k \right)$, the control problem can be formalized as follows.
\vspace{-0.1cm}
\begin{equation}
	\label{eq:wbc}
	\begin{aligned}
		&\exists \ T < \infty,\\ &\text{subject to:} \vspace{-0.05cm}\\
		&\quad \mathbf{x}_0 = \mathbf{x}_{\textnormal{init}}, \ \mathbf{x}_{k+1} = \mathbf{x}_k + f(\mathbf{x}_k, \mathbf{u}_k), \vspace{-0.05cm}\\
		&\quad \mathcal{C} \left( \mathbf{x}_k \right) \in \mathcal{F} \coloneqq \mathcal{F}^g\cup \mathcal{F}^0\cup \mathcal{F}^1, \forall k, \vspace{-0.05cm}\\
		&\quad \mathcal{C} \left( \mathbf{x}_0 \right) \in \mathcal{F}^0, \ \mathcal{C} \left( \mathbf{x}_T \right) \in \mathcal{F}^1,
	\end{aligned}
\vspace{-0.1cm}
\end{equation}
where $f\left( \cdot \right)$ is the system's discrete-time dynamics function and $\{\mathbf{u}_k\}$ is the control sequence as the decision variable.
\vspace{-0.1cm}

\section{Method: Learning Whole-Body Control From Pixels to Action}
\begin{figure} \centering 
	\vspace{0.2cm}
	\includegraphics[width=\columnwidth]{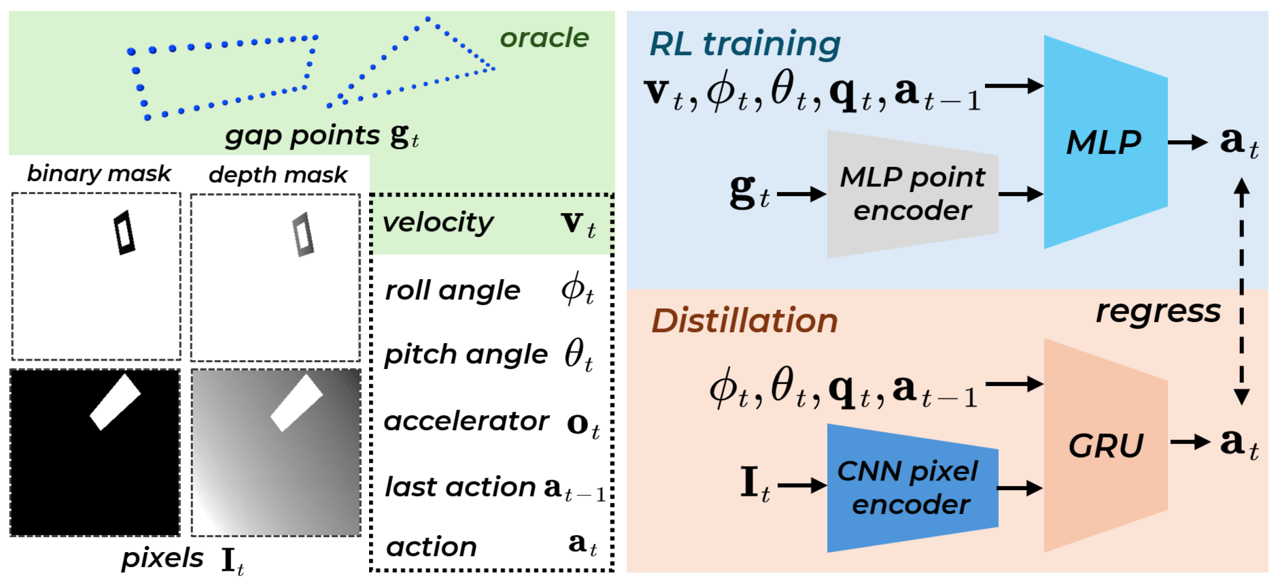} 
	\captionsetup{skip=5pt} 
	\captionsetup{font=footnotesize}  
	\caption{\textbf{The policy learning architecture employed in this paper.} }
	\label{fig:method} \vspace{-0.6cm}     
\end{figure}

\vspace{-0.05cm}
In this section, we present our approach to employ the RL formulation to solve the problem (\ref{eq:wbc}) in a closed-loop manner from sensory observation. The pose and geometry of the gap are prior unknown and not explicitly predicted as interfaces for control. Instead, we design an end-to-end policy directly mapping pixels to the action, as illustrated in Fig. \ref{fig:method}.  \vspace{-0.1cm}   

\subsection{Online Reinforcement Learning from Gap Points}
\vspace{-0.05cm}
\subsubsection{Observation and Action Space}
\begin{wrapfigure}[6]{r}{0.3\textwidth} 
	\vspace{-0.4cm}
	\includegraphics[width=\linewidth]{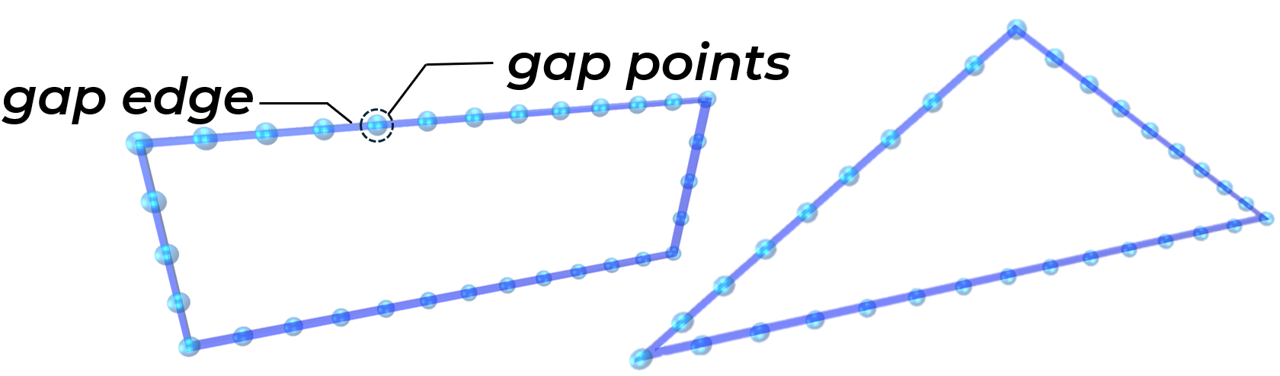}
	\captionsetup{font=footnotesize} 
	\caption{\footnotesize\textbf{Examples of gap points.}}
	\vspace{-0.2cm}
	\label{fig:gap point}
\end{wrapfigure}

In principle, the oracle input should be inferrable from current and historical sensory data. So we choose $n_\mathbf{g}$ ($n_\mathbf{g}$ is 32 in our implementation) 3D points $\mathbf{g}_t$
uniformly sampled along the gap edge, as shown in Fig. \ref{fig:gap point}, and transform it into the body frame  to surrogate the egocentric pixel information. Roll and pitch angles $\mathbf{\phi}_t$ and $\mathbf{\theta}_t$ of the vehicle can be estimated with high precision (within 2 $\deg$ of error) from a single IMU \cite{shalaby2021relative,wu2024scalable}, so we adopt them as input. We also use an oracle information, the body-frame linear velocity $\mathbf{v}_t$.\hspace{-0.3cm} We consider this information to be readily inferrable from the history of IMU readings and actions with little drift in a short term. We also take the accelerometer reading $\mathbf{q}_t$ and the last taken action $\mathbf{a}_{t-1}$ as inputs. Note that we do not include the position as input. The outputs are low-level control commands, collective thrust and body rate, to be executed by a flight controller embedded in the quadrotor.

\subsubsection{Reward Function}
RL optimizes the problem in the form of $
	\max_{\pi} \mathbb{E}_{\mathbf{a}_t\sim \pi \left(  \cdot |\mathbf{o}_t \right)}\left[ \sum\nolimits_t^{}{\gamma ^tr_t} \right]$, where the immediate reward $r_t$ should be designed to find the solution of (\ref{eq:wbc}), such that the policy $\pi$ conditioned on observation $\mathbf{o}_t$ can find the optimal control sequence in a closed loop. 
	\begin{itemize}
		\item  \emph{traversing reward:}
		\vspace{-0.2cm} 
			\begin{align}
				\label{eq:traver reward}
				\mathbb{I} &\left[ \left| x_{t}^{g} \right|\leq l^{\mathcal{C}}
				\text{ and } \left\| \mathbf{p}_t-\mathbf{p}^g \right\| \leq d
				\text{ and } \mathcal{C} \left( \mathbf{x}_k \right) \in \mathcal{F} \right] \cdot \nonumber\\
				&( x_{t}^{g}-x_{t-1}^{g}),
				\vspace{-0.3cm}
			\end{align} 
		where $\mathbb{I} \left[ \cdot \right]$ is the indicator function, $x^{g}$ is the 1-axis coordinate established parallel to the gap normal vector with the origin at the center of the gap's thickness, as illustrated in Fig. \ref{fig:illustration}, $l^\mathcal{C}>0$ is a threshold set according to the size of the collider, $\mathbf{p}^g$ is the geometric center of the gap, and $d>0$ is also a threshold. The indicator approximately determines if the vehicle is traversing the gap at the moment, i.e., $\mathcal{C} \left( \mathbf{x}_k \right) \cap \mathcal{F} _g\ne \varnothing$. Optimizing this term is equivalent to find the solution of (\ref{eq:wbc}).
		\item \emph{shaping reward:} 
		\vspace{-0.2cm}
		\begin{equation}
			\mathbb{I} \left[ \left| x_{t}^{g} \right|>l^{\mathcal{C}} \right] \cdot \left( \left\| \mathbf{p}_{t-1}-\mathbf{p}^g \right\| -\left\| \mathbf{p}_t-\mathbf{p}^g \right\| \right). 
			\vspace{-0.1cm}
		\end{equation}
		This reward, while not always describing the right motion of the quadrotor, helps to alleviate the exploration problem to some extent.  
		\item \emph{jerky motion penalties:} 
		\vspace{-0.2cm}
		\begin{equation}
			-(\lambda _{\mathrm{mag}}\cdot \left\| \left. \mathbf{a}_t \right\| \right. +\lambda _{\mathrm{var}}\cdot \left\| \mathbf{a}_t-\mathbf{a}_{t-1} \right\|), 
			\vspace{-0.2cm}
		\end{equation}
		where $\lambda _{\mathrm{mag}},\lambda _{\mathrm{var}}$ greater than 0 are the penalty coefficients for the magnitude and variation of actions. These penalties are used to encourage a smooth motion. We find this term of reward and tuning the corresponding weight are very crucial for the policy to perform similar smooth motions as model-based methods \cite{wang2022geometrically}.
		\item \emph{aggressiveness constraint:}
		\vspace{-0.1cm}
		\begin{equation}
			\mathbb{I} \left[ \left\| \mathbf{v}_t \right\| \leqslant \mathrm{v}_{\max} \right] \cdot \left( -\exp \left( \left\| \mathbf{v}_t \right\| -\mathrm{v}_{\max} \right) +1 \right).
			\vspace{-0.1cm}
		\end{equation}
		This reward is set to constrain the velocity of the motions.
		\item \emph{distillation-awareness  regularization:} 
		\vspace{-0.1cm}
		\begin{align}
			\hspace{-0.3cm}\lambda _{\mathrm{vis}}\cdot &  \mathbb{I} \left[ \left| x_{t}^{g} \right|>l \right] \cdot \left| \mathbf{g}_{t}^{\mathrm{vis}} \right|- 
			\lambda _{\boldsymbol{\nu }}\cdot \mathbb{I} \left[ \left| x_{t}^{g} \right|>l \right] \cdot \left\| \boldsymbol{\nu }_{t}^{x}-\boldsymbol{\nu }_{t}^{g} \right\| \\ -\nonumber
			&\lambda _{\mathbf{n}}\cdot \mathbb{I} \left[ \left| x_{t}^{g} \right|\leq l \right] \cdot \left\| \boldsymbol{\nu }_{t}^{x}-\mathbf{n}^g \right\|,
			\vspace{-0.1cm}
		\end{align}
		 where $\left| \mathbf{g}_{t}^{\mathrm{vis}} \right|$ is the number of visible gap points, i.e., the gap points in the limited field of view (FOV), and $\boldsymbol{\nu }_{t}^{x}$, $\boldsymbol{\nu }_{t}^{g}$ and $\mathbf{n}^g$ are the direction of body x-axis of the quadrotor (the directed centerline of FOV), the direction pointing from the body to the gap center, and the normal of the gap plane, respectively. We call the rewards above as distillation-awareness reward since it helps to \emph{reduce the discrepancy between the gap point and pixel observations}. Specially, the first two terms encourage actions improving visibility of the whole gap, and  the last term regularizes the behavior of the quadrotor when it is approaching the gap and the gap becomes invisible. 
		 
	\end{itemize}

We note the weight of the traversing reward is significantly larger than those of others.

\subsubsection{Termination Condition}
An episode terminates when one of the following happens: (i) $\mathcal{C} \left( \mathbf{x}_t \right) \in \mathcal{F}^1$, i.e., the quadrotor has crossed the gap, (ii) the agent is out of an artificially defined world box, (iii) the quadrotor collides with the gap plane, and (iv) the episode steps achieves a certain number. 
\subsubsection{Policy Representation}
The policy is represented by a neural network consisting of a simple gap point encoder and a feedforward output network. The point encoder receiving the gap points as input is a multiple layer (MLP) with an intermediate global max-pooling layer to encode features that are invariant to the order of points \cite{qi2017pointnet}. The feedforward network is an MLP that fuses the point features and other observations to output actions. \vspace{-0.2cm}

\begin{figure}[ht]
	\centering
	\begin{minipage}[h]{0.24\textwidth}
		\vspace{-2.5cm}
  		\includegraphics[width=\textwidth]{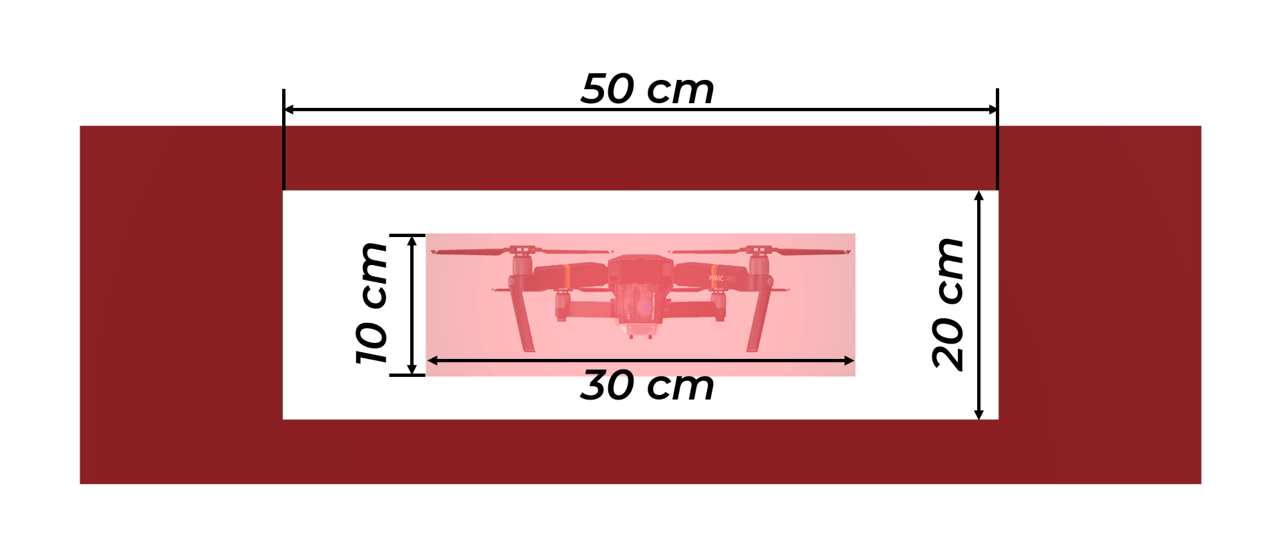}
	\end{minipage}
	\begin{minipage}[b]{0.24\textwidth}
		\centering
		\includegraphics[width=\textwidth]{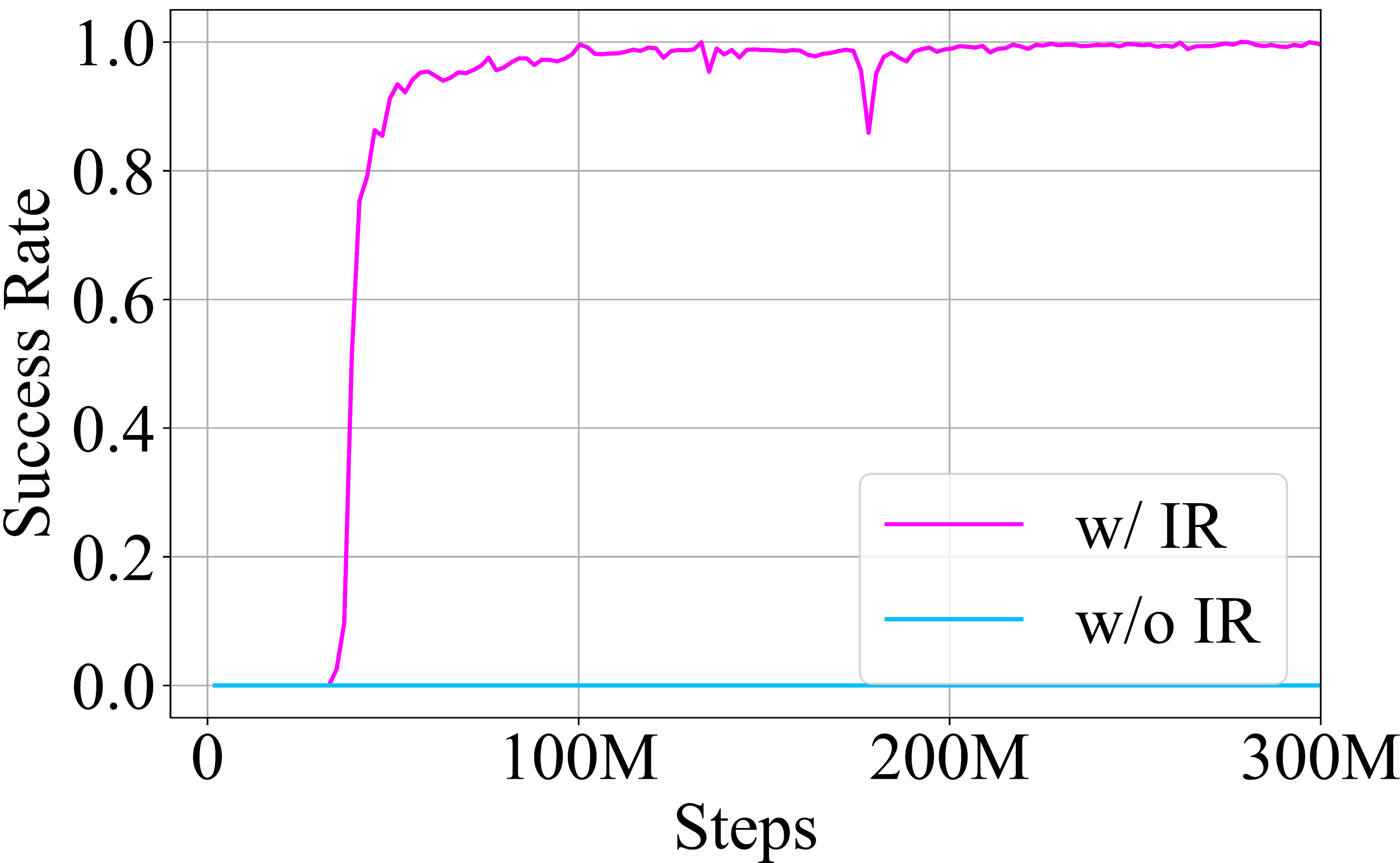}
	\end{minipage}
\captionsetup{font=footnotesize} 
	\caption{\textbf{A motivating example of IR.} The left figure is a visualization of the sizes of the collider vs. the gap. The right figure is the success rate evolution.}
	\label{fig:example}
	\vspace{-0.4cm}
\end{figure}

\subsubsection{Informed Reset}
We begin the introduction of our reset scheme, called the informed reset (IR),  with a motivating example in Fig. \ref{fig:example}. In this example, the collider of the quadrotor is modeled as a cuboid and is trained to fly through a rectangular gap tilted so that the long side is \emph{vertical} to the ground. From the results in Fig. \ref{fig:example}, the policy is allowed for rapid mastery of this flight skill once the initial success occurs in the IR-enabled case, whereas directly training the policy in such a challenging task leads to an exploration failure.

The pipeline of IR is as follows. First, we define the task distribution at triple levels: the gap geometry, the gap pose, and the initial state of the quadrotor relative to the gap. We divide each level into subspaces, e.g., we have $i$ kinds of gap, divide the pose into $j$ disjoint sets, and divide the initial state into $k$ disjoint sets, thus obtaining $i\times j\times k$ task subspaces. 

Then, we follow the quotient space-based method for $\mathrm{SE}\left( 3 \right)$ trajectory optimization in our previous work \cite{wang2022geometrically}, where the gap is represented by polytopes, to generate one trajectory for each task space. These trajectories are generated efficiently compared to other model-based methods thanks to the advanced computational techniques proposed in \cite{wang2022geometrically}. In aggregate, $i\times j\times k $ trajectories are collected offline.

We propose to reset the agent in the state randomly sampled from the corresponding trajectories of the task subspace, which includes the position, attitude, velocity and acceleration. We set a 50\% probability of sampling the states on these trajectories, and the rest follows a randomized sampling of initial states. States that satisfy the condition of reward (\ref{eq:traver reward}) are not sampled. The yaw angle of the reset state is imposed to let the quadrotor be visable to the target. This method shares a similar spirit with those in \cite{salimans2018learning, peng2018deepmimic, uchendu2023jump}, which use informed states to reset the agent. The idea behind this reset scheme is logically intuitive: the high-quality trajectories impose the exploration distribution around some crucial states and starting the episode from these states can make random exploration actions more likely to be highly rewarded. These beneficial actions are \emph{reinforced} and more likely to be followed in the future, which gradually biases the explored distribution. 

\begin{figure*}[ht]
	\vspace{-0.3cm}
	\centering
	\includegraphics[width=0.48\textwidth]{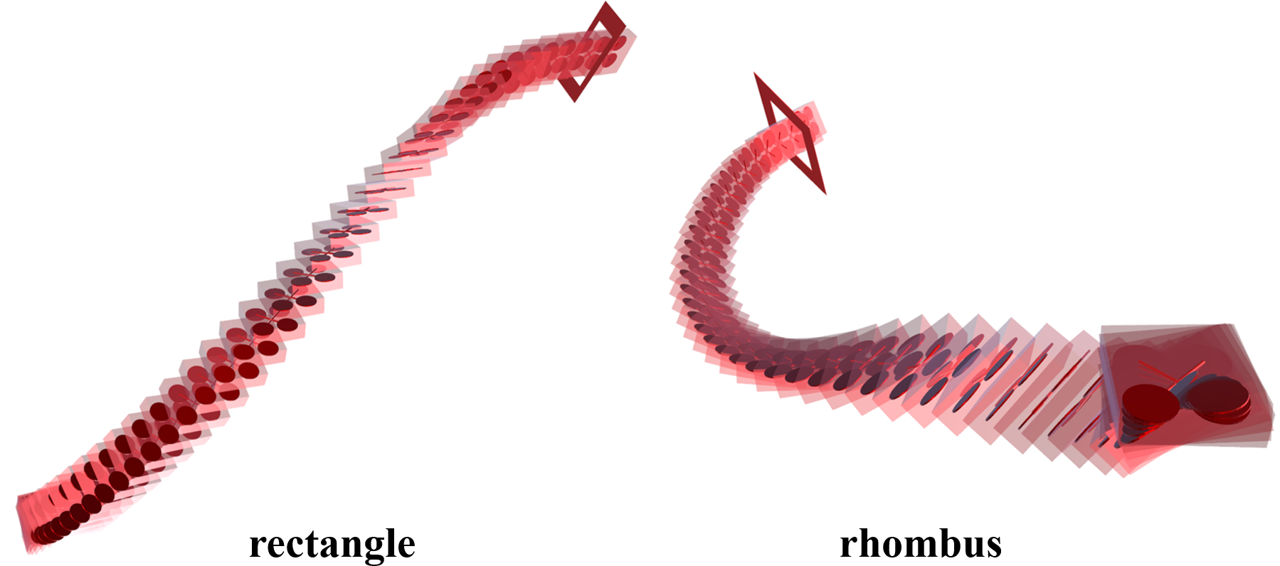}
	\includegraphics[width=0.48\textwidth]{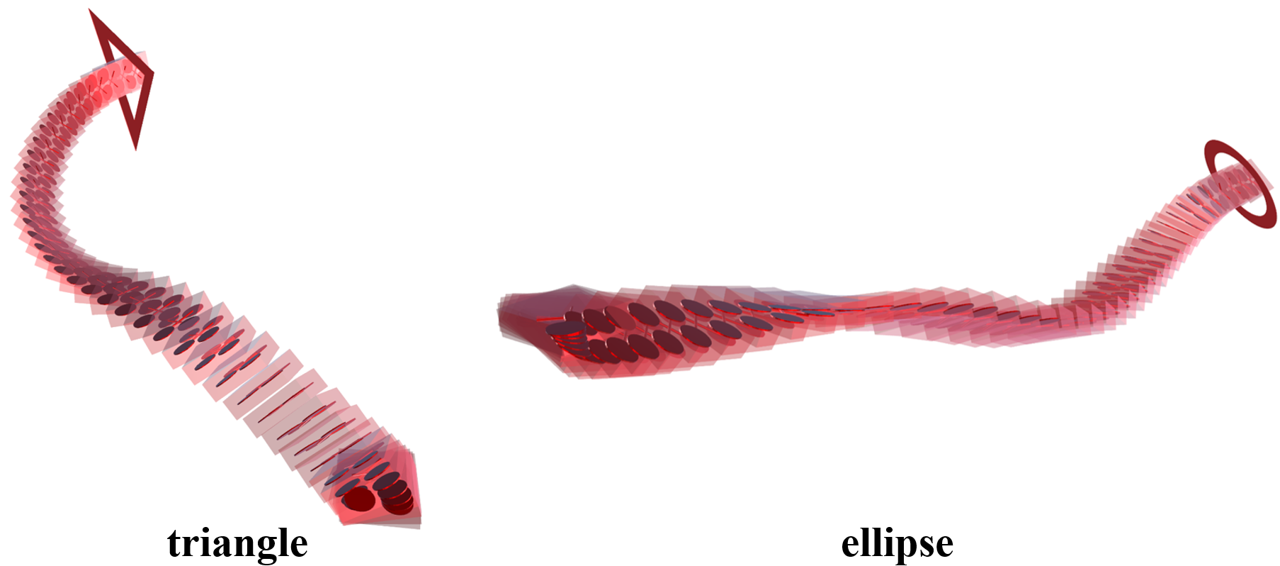}
	\captionsetup{font=footnotesize} 
	\caption{\textbf{The visualization of the policy rollouts to drive the quadrotor through various gaps in the \emph{'hard'} setup.}}
	\label{fig:rollouts}
	\vspace{-0.3cm}
\end{figure*}

\begin{figure*}[ht]
	\centering
	\begin{minipage}[b]{0.248\textwidth}
		\centering
		\includegraphics[width=\textwidth]{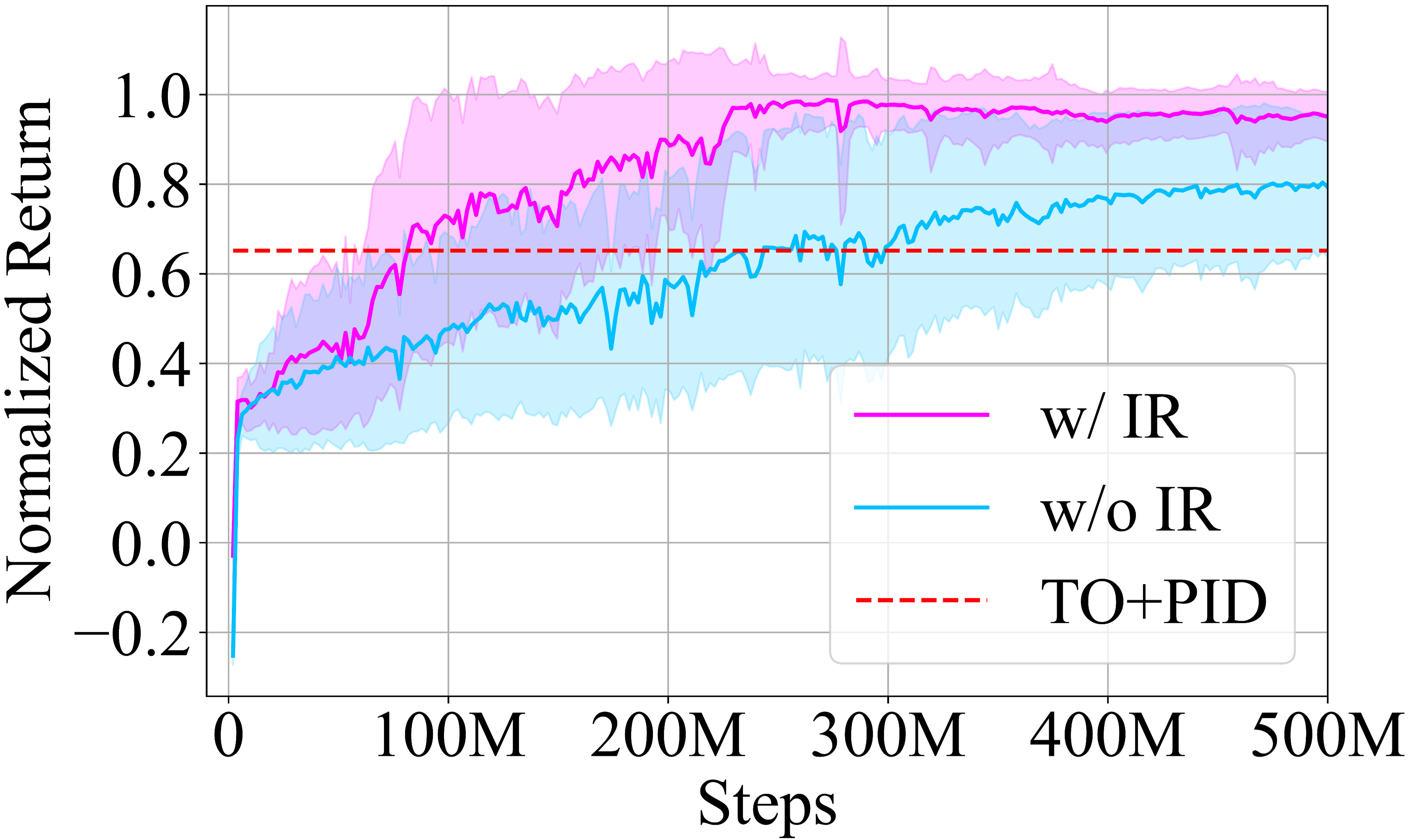}
	\end{minipage}
	\hfill
	\begin{minipage}[b]{0.24\textwidth}
		\centering
		\includegraphics[width=\textwidth]{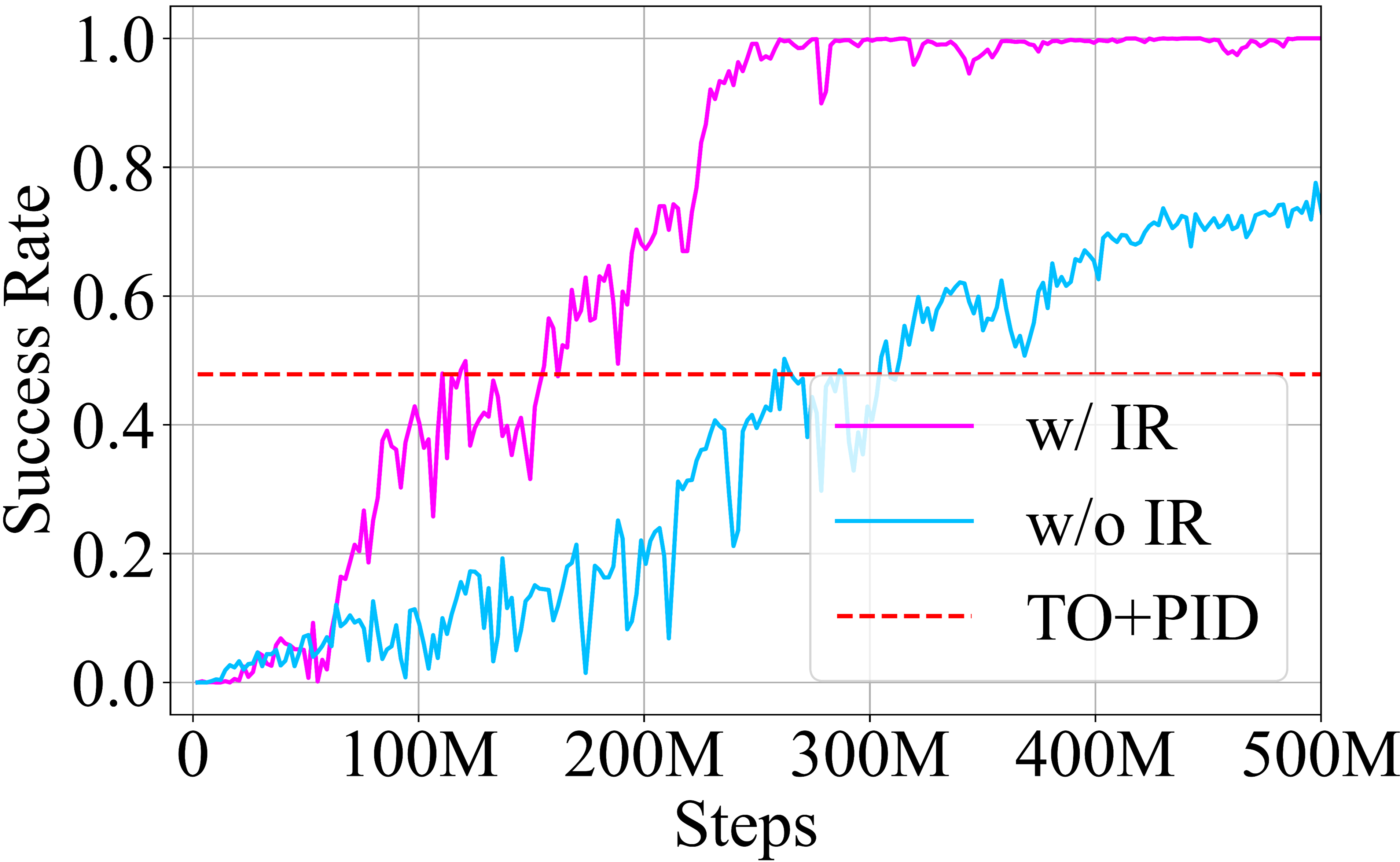}
	\end{minipage}
	\hfill
	\begin{minipage}[b]{0.248\textwidth}
		\centering
		\includegraphics[width=\textwidth]{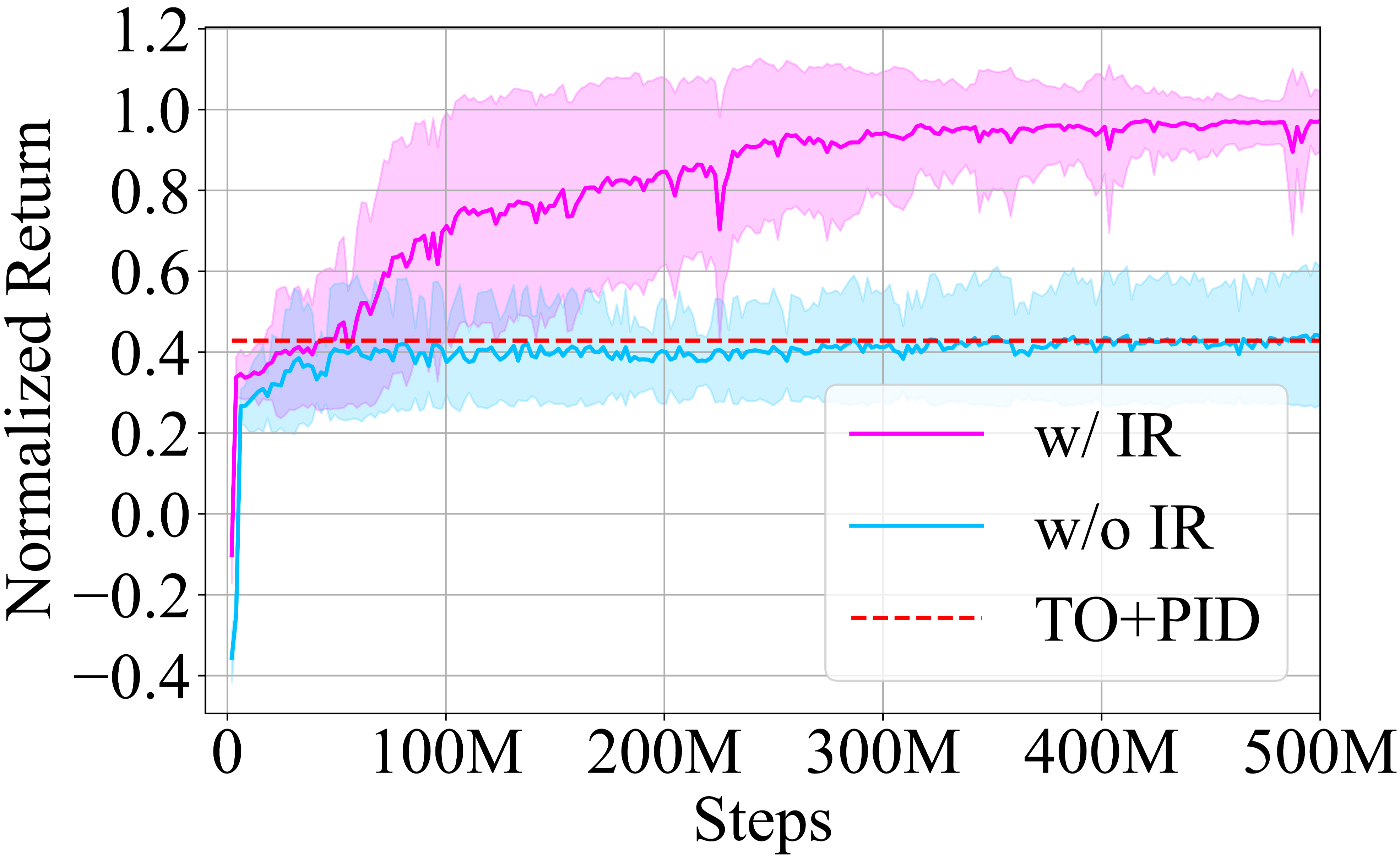}
	\end{minipage}
	\hfill
	\begin{minipage}[b]{0.242\textwidth}
		\centering
		\includegraphics[width=\textwidth]{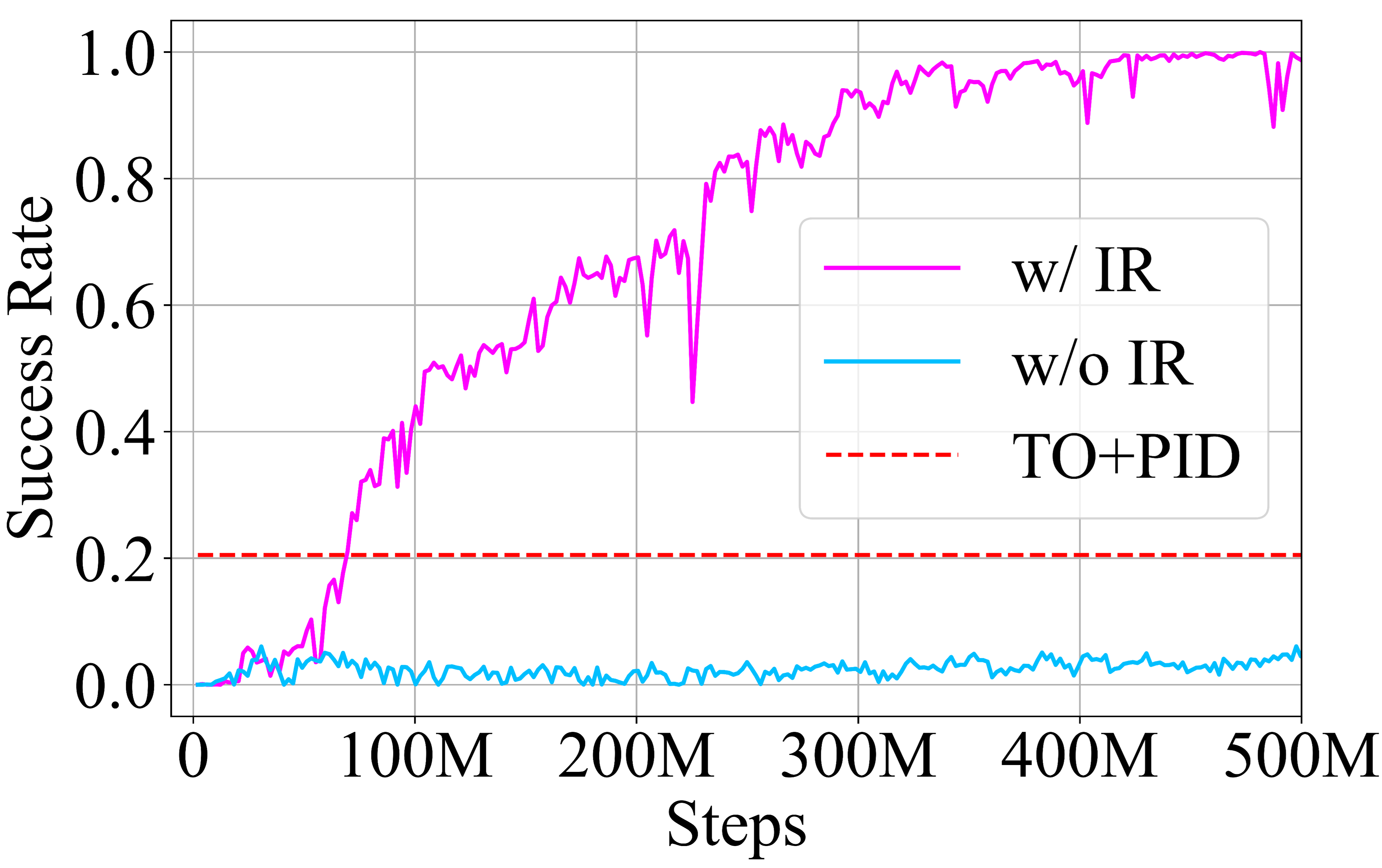}
	\end{minipage}
	\captionsetup{font=footnotesize} 
	\caption{\textbf{The undiscounted normalized return and success rate of our method and baselines at two different difficulty levels.} The left two figures are for 'simple' setup and the right two are for 'hard'. 'TO+PID' in the figure is the baseline performance from model-based techniques described in Sec. \ref{sec:implementation}.}
	\label{fig:results}
	\vspace{-0.6cm}
\end{figure*}

\subsection{Online Distillation via Supervised Learning}
\label{sec:distillation}
A binary mask of the gap instance and the corresponding depth image, as in Fig. \ref{fig:method}, are the pixel inputs of the final policy, similar to those in \cite{liu2024visual}. The utilization of such preprocessed information is motivated by three ways: (i) it reduces complexity in raw visual input, (ii) it is easily accessible adopting the recent results from zero-shot segmentation model \cite{nanosam2023, ravi2024sam} without a specialized detector (e.g., in \cite{geles2024demonstrating}), and (iii) the sim-to-real gap is relatively small. 

The current pixel input $\mathbf{I}_t$ are encoded using a lightweight convolutional network and the obtained visual feature is concatenated with $\mathbf{\phi}_t$, $\mathbf{\theta}_t$, $\mathbf{o}_t$ and $\mathbf{a}_{t-1}$ to input into a one-layer recurrent network, gated recurrent unit (GRU), to maintain a memory. A following feedforward MLP outputs the action.

For stable training, we first collect data by rolling out the RL policy and perform offline behavior cloning using the output of the RL policy as supervision to pre-train the pixel-based policy. Then we use the dataset aggregation (DAgger) \cite{ross2011reduction} as a meta-algorithm and the same behavior cloning loss to train the pixel-based policy. DAgger is an online imitation learning method that effectively suppresses the covariate shift problem in both theory \cite{ross2011reduction} and practice \cite{pan2017agile}.

We note that the problem definition we propose in Sec. \ref{sec:problem} is adaptable to various environments. For example, we can choose two tree trunks as mask, define gaps with thickness between two trunks and set the gap to a near height to that of the initial position. In this way, we can successfully transfer knowledge for this landscape (see Fig. \ref{fig:head}), thanks to the flexibility of distillation-fashion methods. The assumption therein is that the cues for taking the action can be inferred from historical sensory data.

\section{Evaluation}
\vspace{-0.1cm}
In this section, we conduct experiments in simulation and demonstrate that
\begin{itemize}
	 \item our method achieves agile and smooth maneuver through very narrow gaps, as shown in Fig. \ref{fig:rollouts};
	 \item IR contributes to efficient online RL and alleviates  the exploration challenge in tasks that are hard to solve with random initialization;
	 \item specialized reward design is necessary to suppress the performance degradation during distillation;
	 \item distillation is a better choice than direct pixel-based RL. 
\end{itemize}

\subsection{Implementation and Setup}
\label{sec:implementation}
The RL policy is trained in a customized simulator developed based on the open-source Flightmare \cite{song2021flightmare}, where we use a single Intel i9-14900K centralized processing unit (CPU) to simulate the dynamics transitions, and an NVIDIA RTX 4090 graphic processing unit (GPU) for update and inference of neural networks and image rendering. We simulate 2048 environments to collect data for online RL and use proximal policy optimization (PPO) \cite{schulman2017proximal} algorithm that is demonstrated to be powerful with a large throughput of data. 

The collider of the quadrotor is modeled as a 30 $\mathrm{cm} \times$30 $\mathrm{cm}\times$10 $\mathrm{cm}$ cuboid. The quadrotor is equipped with a stereo fisheye camera with a 120 $\deg$ depth prediction  (such as \cite{liu2024omninxt}) to alleviate issues caused by limited FOV. The image input is 128 $\times$ 128. We clip the output thrust to keep the thrust-to-weight ratio in the range of 0.56 to 2.8 and the body rate of each axis in the vehicle's body is clipped under 5 $\mathrm{rad/s}$. 

The command constraints in the trajectory planner used in IR and benchmark results in Fig. \ref{fig:results} are consistent with those of RL. The low-level controller is implemented as a PID controller. We manually tune the parameters of TO and PID. Using a CPU parallelization, we can collect over 200 complete trajectories, around 12000 control steps, per second. Note that while this is considerably fast compared to other model-based approaches, it is only about 1/50th of the sampling rate in RL (which includes network inference). We collect around 8000 trajectories for our task distribution.

\begin{figure*}[t]
	\vspace{0.1cm}
	\centering
	\setlength{\tabcolsep}{5pt}
	
	\hspace{-0.3cm}
	\begin{minipage}[l]{0.65\textwidth} 
		\centering
		\renewcommand{\arraystretch}{1.1}
		\vspace{-4.7cm}
		\begin{footnotesize}
			\begin{tabular}{c c c}
				\toprule
				\multirow{2}{*}{Geometry} & \multicolumn{2}{c}{Vertices ($\mathrm{cm}$)}  \\ \cline{2-3} 
				& Simple & Hard \\ \midrule
				\textbf{rectangle} & $(25, 10), (-25, 10), (-25, -10), (25, -10)$ & $(22, 8), (-22, 8), (-22, -8), (22, -8)$ \\
				\textbf{rhombus} & $(42, 0), (0, 12), (-42, 0), (0, -12)$ & $(40, 0), (0, 10), (-40, 0), (0, -10)$ \\
				\textbf{triangle} & $(24\sqrt{3}, 0), (0, 24), (-24\sqrt{3}, 0)$ & $(22\sqrt{3}, 0), (0, 22), (-22\sqrt{3}, 0)$ \\
				\textbf{ellipse} & $(30, 0), (0, 10), (-30, 0), (0, -10)$ & $(30, 0), (0, 10), (-30, 0), (0, -10)$ \\
				\textbf{parallellogram} & $(18, 0), (-18, 26), (-18, 0), (18, -26)$ & $(16, 0), (-16, 24), (-16, 0), (16, -24)$ \\
				\bottomrule
			\end{tabular}
		\end{footnotesize}
		\captionsetup{font=footnotesize,justification=centering}
		\captionof{table}{\textbf{Geometry and sizes of different gaps expressed in 2D vertices coordinates.}} 
		\label{tab:geometry}
	\end{minipage}
	\hspace{0.75cm}
	\begin{minipage}[b]{0.29\textwidth} 
		\centering
		\includegraphics[width=1\textwidth]{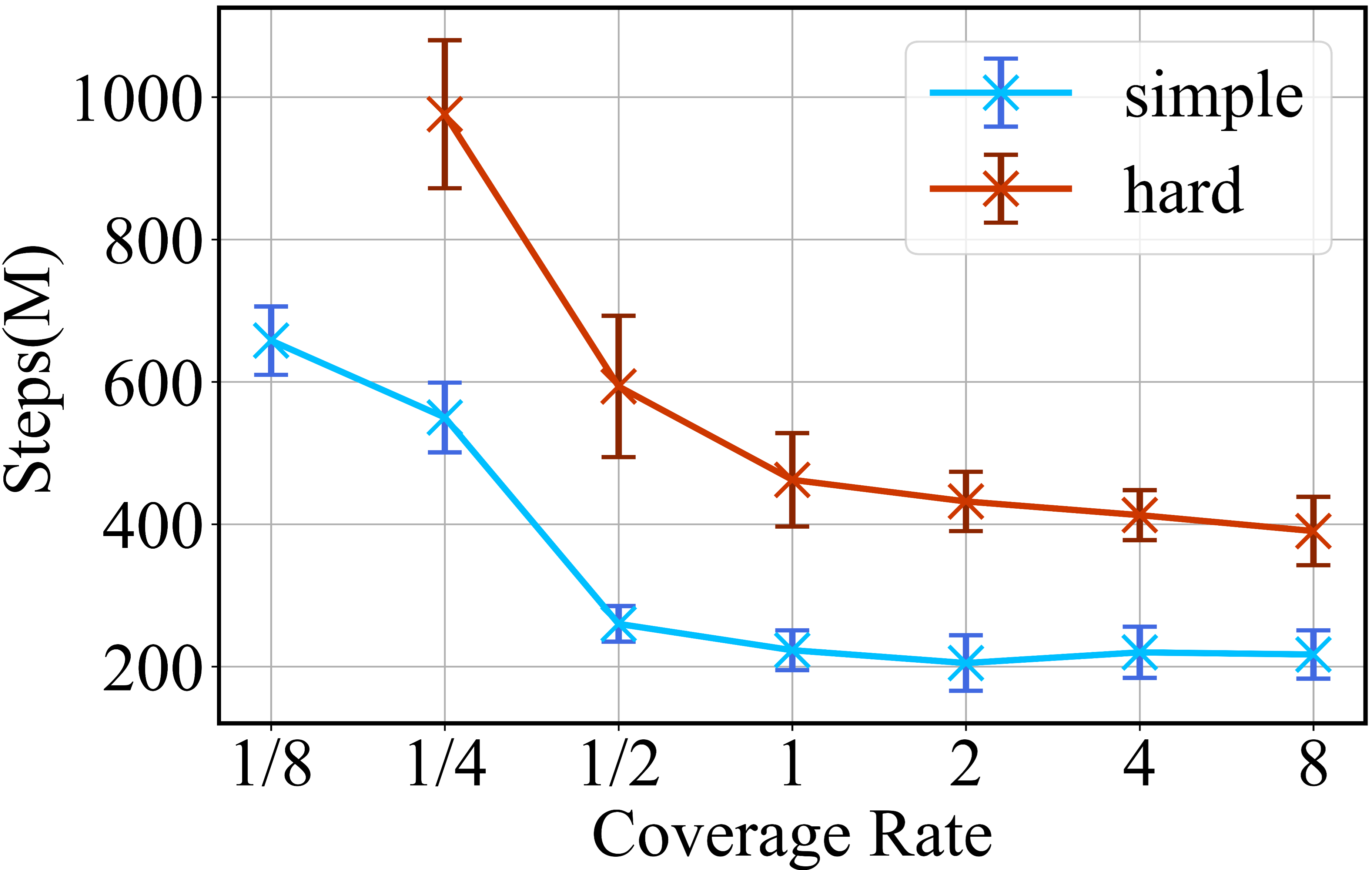}
		\captionsetup{font=footnotesize,justification=centering}
		\captionof{figure}{\textbf{Efficiency enhancement of different coverages.}} 
		\label{fig:coverage}
	\end{minipage}
\vspace{-0.7cm}
\end{figure*}

\subsection{Baseline and Ablation Studies of Online RL}
\vspace{-0.1cm}
\label{sec:rl results}
We design two tasks including 5 different gaps, respectively, as illustrated in Tab. \ref{tab:geometry}  with different levels of difficulty.  In the 'hard' setup, the area of the gap is almost less than twice the cross-section area of the quadrotor's collider (e.g., for \textbf{rectangle}, this ratio is 1.96). Even the 'simple' setup is more challenging than the simulation results in previous work with the quadrotor's full state as input \cite{xiao2021flying} and fixed gap geometry. In both setups, the roll angle of the gaps and initial positions of the quadrotor is randomized. We can see from Fig. \ref{fig:results} that the naive PPO method can achieve a considerable success rate in the 'simple' setup. We attribute this to the capability of capitalizing large throughput of data. 

As can be seen from the results,  in the 'simple' setup, training with IR takes much less data than the naive PPO baseline to achieve a near-perfect performance. It is more dramatic to see that the naive PPO fails at exploring to a successful action sequence to fly through most narrow gaps under the 'hard' setup. We hypothesize that this difficulty arises from the fact that (i) the gap is too small for a naturally explored sequence of actions that can get at substantial rewards (\ref{eq:traver reward}) when approaching the gaps, and, at the same time, (ii) the reward components that encourage smooth motions or visibility, which importantly contribute to the desired flight behavior, hinder attempts from aggressive intention to change the vehicle's attitude. As a result, the policy learned with naive PPO is trapped in a lazy suboptimal solution.

We also conduct a parameter experiment on the effect of the coverage of trajectories used in IR over the entire task distribution on the advantages it brings.  We consider $\sim$8000 trajectories used in the previous experiments as the baseline setup and the coverage rate is defined as the ratio of the number of trajectories, which also uniformly cover the task distribution, to that of the baseline setup. We report the steps consumed to achieve a 99\% success rate, as shown in Fig. \ref{fig:coverage}. It can be seen that when fewer trajectories are used, the increased coverage significantly improves the efficiency of online RL, while when the number continues to increase, the improvement in efficiency reaches saturation. We therefore recommend using, e.g., bisection method, to quickly find the saturation interval and then conducting subsequent training.

\subsection{Baseline and Ablation Studies of Observation Distillation}
In this subsection we show pixel-based policies under the task settings in Sec. \ref{sec:rl results} obtained via distillation. The baseline is the one that uses RL to directly learn from pixels and uses the same GRU architecture with IR enabled, of which the training time is \emph{the sum of the two phases} of the distillation-based method. The rate of image rendering is around 1/100th of that of using the gap point. 

As expected, although we add a memory module in the pixel-based policy to make the information contained in the gap points used in RL inferrable from the limited FOV vision, if we do not force this intention in the reward, the performance degradation of distillation becomes relatively significant. Tab. \ref{tab:distillation results} shows the relevant results. 

Unfortunately, while a small performance degradation can be achieved using visibility rewards under 'simple' setup, we can see that there is still a significant performance loss under 'hard' setup from the near-perfect performance of its RL counterpart. One way to mitigate this loss is distilling using a larger task variation instead of only the 'hard' one, i.e., we distill two RL policies obtained by training under the two setups into a unified pixel-based policy. The relevant results are reported in the 'Simple+Hard' column of Tab. \ref{tab:distillation results}, where the statistics are  \emph{only under the 'hard' setup}. 

From the results we can see that the method of learning directly from RL suffers from inefficiencies at lower rates of data generation and when using relatively complex network structures. By comparision, supervised learning with first-order gradient information can easily utilize higher dimensional inputs as well as complex networks.

\setlength{\tabcolsep}{5pt} 
\begin{table}[ht]
	\vspace{0.1cm}
	\centering
	\renewcommand{\arraystretch}{1.15} 
	\setlength{\tabcolsep}{8pt}
	\begin{tabular}{c c c c c c c}
		\toprule
		\multirow{2}{*}{Method} & \multicolumn{2}{c}{Simple} & \multicolumn{2}{c}{Hard} & \multicolumn{2}{c}{Simple+Hard} \\ \cline{2-7} 
		& SR (\%) & Ret & SR (\%) & Ret & SR (\%) & Ret \\ \midrule
		\textcolor{magenta}{w/ DA} & 97.6 & 0.93 & 93.6 & 0.91 & 95.7 & 0.92 \\
		w/o DA & 89.1 & 0.90 & 84.7 & 0.86 & 82.5 & 0.82 \\
		RL & 5.2 & 0.25 & 1.8 & 0.12 & 1.5 & 0.15 \\
		\bottomrule
	\end{tabular}
	\captionsetup{font=footnotesize,justification=centering}
	\caption{\textbf{Results of baseline and ablation for distillation.} SR means success rate, Ret means (normalized) return, and DA means the distillation-awareness reward. The \textcolor{magenta}{meganta} text indicate ours method.}
	\label{tab:distillation results} 
	\vspace{-0.5cm}
\end{table}

\section{Discussions and Outlook}
\subsubsection{Potential and challenges of sim-to-real transformation}
All the inputs to the pixel-based policy in this work can be obtained from onboard sensors. In particular, the choice of the pixel input should demonstrate a relatively small sim-to-real gap compared to other visual input modes. Therefore, the current design of the observation space is promising for deployment in the real world.

However, depth sensing can be noisy, whereas the binary mask, by contrast, offers pixel-level accuracy. Nevertheless, depth information ensures that the geometry and orientation of the gaps are observable in the initial stages of flight, which is necessary for flight through gaps with diverse geometries. We need additional design, such as preprocessing depth images to add various types of noise, to encourage the pixel-based policy to implicitly combine and switch depth versus mask information through data fusion, e.g., the historical actions and proprioception, for a precise flight. 

\subsubsection{Scaling up the task variation}
A nice vision is to learn a versatile policy enabling flight through a wide variety of gaps, finding the most cost-effective but safe way to traverse through them. Such a policy can be useful in practical to enable a fly robot to travel through narrow space in artificial and natural landscapes in the real world, where the gaps are often irregularly shaped. This requires a versatile policy trained under gaps with various geometries and sizes so that it generalizes to gaps unseen during training.

\subsubsection{A simple method drawing on the knowledge from model-based techniques}
Common ways of combining model-based techniques and data-driven approaches include distilling knowledge from model-based approaches into neural networks \cite{pan2017agile}, using a hierarchical  design \cite{zhao2024learning}, and learning dynamics models \cite{o2022neural}, etc. In contrast to these approaches, this work provides a simple method to utilize the knowledge of model-based methods while minimally restricting the exploration of RL. We hope that this approach can be demonstrated successfully in more applications.

\bibliographystyle{plainnat}
\bibliography{references}
\end{document}